\documentclass[runningheads]{llncs}
\usepackage{amsfonts}
\usepackage{graphicx}
\usepackage{subcaption}
\usepackage{hyperref}
\usepackage{comment}
\usepackage{xcolor}
\usepackage{tabu}
\usepackage{multirow}
\usepackage{amsmath}
\usepackage[skip=3pt]{caption}
\usepackage{tabu}
\usepackage{array}
\begin{document}
\title{
Enhancing Gait Video Analysis in Neurodegenerative Diseases by Knowledge Augmentation in Vision Language Model
\thanks{Supported by French National project ``ANR ArtIC: AI for Care'', French Minister.}
} 
\titlerunning{Enhancing GVA in Neurodegenerative Diseases by KA in VLM}
\authorrunning{D. Wang et al.}
\author{Diwei Wang\inst{1}
\and
Kun Yuan\inst{1} \and Candice Muller\inst{2} \and Frédéric Blanc\inst{1,2}
\and Nicolas Padoy\inst{1} \and Hyewon Seo \inst{1}
}
\institute{ICube laboratory, University of Strasbourg, CNRS, France
\\
\email{\{d.wang, kyuan, f.blanc, npadoy, seo\}@unistra.fr}\\
\and Hôpital de la Robertsau, France \\
\email{\{candice.muller, frederic.blanc\}@chru-strasbourg.fr}}

\maketitle              
%
\begin{abstract}
We present a knowledge augmentation strategy for assessing the diagnostic groups and gait impairment from monocular gait videos. Based on a large-scale pre-trained Vision Language Model (VLM), our model learns and improves visual, textual, and numerical representations of patient gait videos, through a collective learning across three distinct modalities: gait videos, class-specific descriptions, and numerical gait parameters. Our specific contributions are two-fold: First, we adopt a knowledge-aware prompt tuning strategy to utilize the class-specific medical description in guiding the text prompt learning. Second, we integrate the paired gait parameters in the form of numerical texts to enhance the numeracy of the textual representation. Results demonstrate that our model not only significantly outperforms state-of-the-art methods in video-based classification tasks but also adeptly decodes the learned class-specific text features into natural language descriptions using the vocabulary of quantitative gait parameters. The code and the model will be made available at our project page: \url{https://lisqzqng.github.io/GaitAnalysisVLM/}.

\end{abstract}
\vspace{-3mm}
\section{Introduction}
\vspace{-1mm}
While quantitative gait impairment analysis has proven to be an established method for accessing neurodegenerative diseases and gauging their severity \cite{pathologicalSignature,QuantativeGA,ImpactofEnvironment,muller2018correlation}, current clinical assessments are used in highly restricted contexts, posing significant challenges: Not only do they often require specialized equipment, such as force plates or IMU sensors, but they also struggle to capture moments with prominent symptoms during clinical visits, which are somewhat special occasions for patients. Analysing motor symptoms from video offers new possibilities, enabling cost-effective monitoring, remote surveillance without the need of frequent in-person clinic visits, thereby facilitating timely and personalized  assessment. \\
\indent Naturally, there have been recent efforts to develop a single 2D-RGB-camera-based gait analysis system, with the majority leveraging advancements in deep learning. Albuquerque et al \cite{St-LSTMGaitClassify} develop a spatiotemporal deep learning approach by producing a gait representation that combines image features extracted by Convolutional Neural Networks (CNNs), chained with a temporal encoding based on a LSTM (Long Short Term Memory) network. Sabo et al \cite{sabo2022estimating} have shown that Spatiotemporal-Graph Convolution Network models operating on 3D joint trajectories outperform earlier models. In the work by Lu et al \cite{lu2020miccai}, 3D body mesh and pose are extracted and tracked from video frames, and the sequence of 3D poses is classified based on MDS-UPDRS gait scores \cite{mds-updrs2008} using a temporal CNN. Wang et al \cite{max-gr2023} have developed a dedicated 3D skeleton reconstructor tailored for gait motion, incorporating a gait parameter estimator from videos and a multihead attention Transformer for similar classification tasks. Among  methods for non-pathological gait analysis, GaitBase \cite{opengait2023}
combines improved spatial feature extraction and temporal gait modeling for appearance-based gait recognition, in both indoor and outdoor settings. \\
\indent Existing works face challenges in handling insufficient pathological gait data and imbalances with normal data, promoting strategies such as a self-supervised pretraining stage prior to the task-specific supervision \cite{sabo2022estimating}, or the employment of crafted loss functions \cite{lu2020miccai}. Nevertheless, the need for data-efficient approaches with superior performance is crucial in video-based pathological gait classification. Meanwhile, the recent emergence of large-scale pre-trained vision-language models (VLMs) has demonstrated remarkable performance and transferability to different types of visual recognition tasks~\cite{clip2021,miech2020end}, thanks to their generalizable visual and textual representations of natural concepts. 
In the context of medical image analysis, VLMs tailored to various medical imaging tasks via finetuning~\cite{huang2023visual}, multimodal global and local representation learning~\cite{huang2021gloria}, knowledge-based prompt learning~\cite{qin2022medical,kapt2023}, knowledge-based contrastive learning on decoupled image and text modality~\cite{wang2022medclip}, and large-scale noisy video-text pretraining~\cite{yuan2023learning}.\\
\indent Inspired by these works, we propose a new approach to transfer and improve representations of VLMs for the pathological gait classification task in neurodegenerative diseases. Concretely, we model the prompt's context with learnable vectors, which is initialized with domain-specific knowledge. Additionally, numerical gait parameters paired with videos are encoded and aligned with the text representation with a contrastive learning. 
During training, the model learns visual and text representations capable of understanding both the class-discriminating 
and numerical features of gait videos. To our knowledge, our work represents the first attempt to deploy VLM for the analysis of pathological videos.

\vspace{-3mm}
\section{Method}
\begin{figure}[!ht]
    \centering
\includegraphics[width=\textwidth]{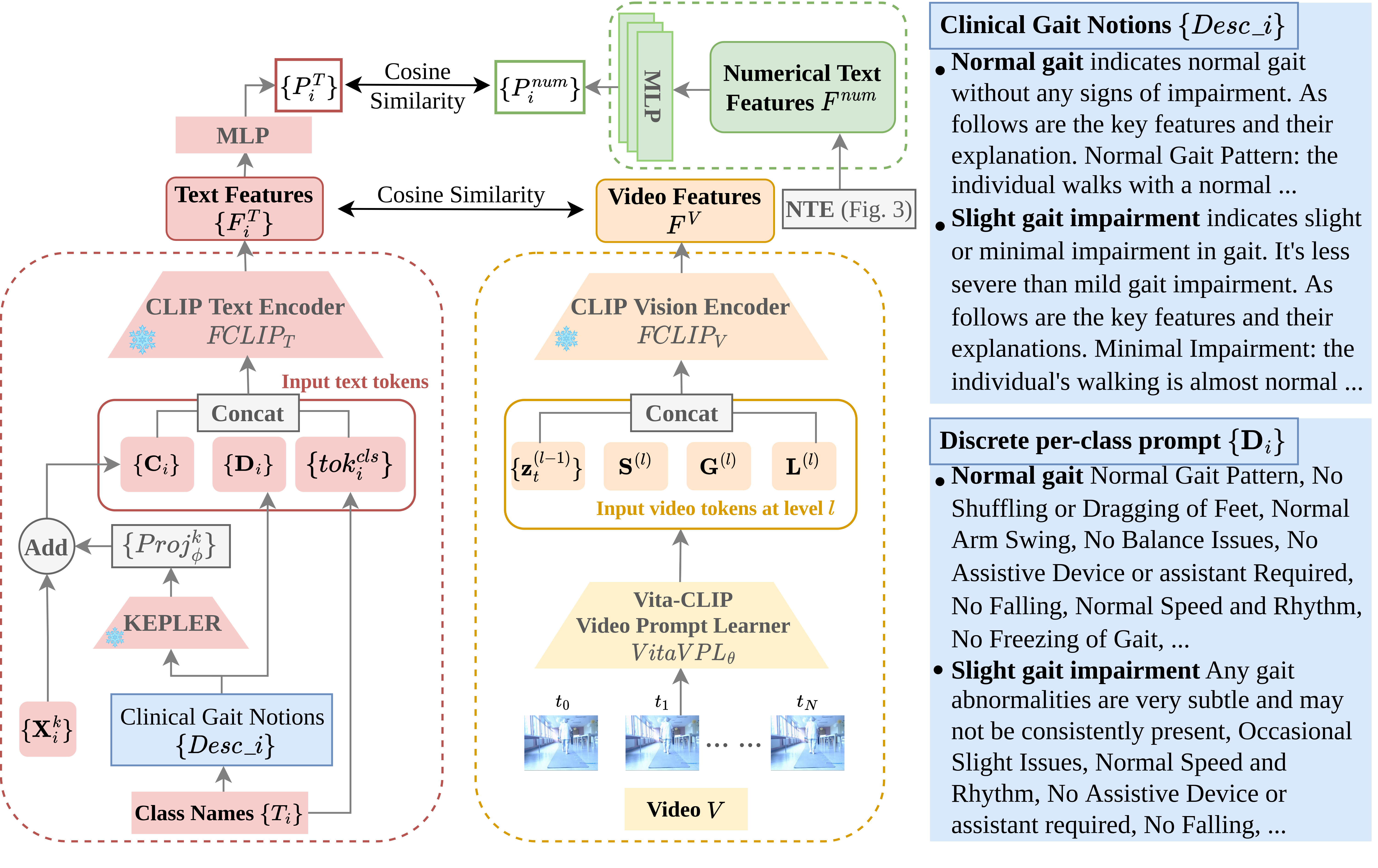}
    \caption{Overview of our cross-modality model for video-based clinical gait analysis (left), alongside clinical gait notions and per-class descriptions of gait classes utilized for prompt initialization (right). Three colored blocks represent the text- and video encoding pipelines, and the text embedding of numerical gait parameters, respectively.}
    \label{fig: architecture}
\vspace{-8pt}
\end{figure}

\vspace{-1mm}
An overview of our method is shown in Fig.\ref{fig: architecture}. We utilize three distinct modalities to enhance the accuracy and the reliability of the VLM in classifying medical concepts: gait videos, class-specific medical descriptions and numerical gait parameters. Our knowledge augmentation strategy consists of two parts: First, we adopt a knowledge-aware prompt learning strategy to exploit class-specific description in the text prompts generation, while leveraging the pre-aligned video-text latent space (Sec.\ref{sec: kapt}).
Second, we incorporate the associated numerical gait parameters as numerical texts to enhance the numeracy within the latent space of the text (Sec.\ref{sec: num gait text}). Once the training is complete, we run a classifier solely on the video features extracted from the input video.

\vspace{-3mm}
\subsection{Dataset and preprocessing}
\label{sec:dataset preprocessing}

\textbf{Dataset.} Our study leverages a dataset comprising 90 gait videos from 40 patients diagnosed with neurodegenerative disorders and 3 healthy controls, as detailed in \cite{max-gr2023}. 
Moreover, 28 gait video clips featuring healthy elderly individuals have been added, chosen from the TOAW archive \cite{toaw2022} based on specific criteria (Berg Balance Scale $\ge{45}$, 0-falls during last 6 months, etc.), bringing the total number to 118 clips. All the videos are recorded at 30 fps, each capturing a one-way walking path of an individual. The patients 
were instructed to walk forth and back on a GAITRite (\url{https://www.gaitrite.com/}) pressure-sensitive walkway, providing a set of gait parameters as outlined in Table \ref{tab: 29 gait params} in Supplementary Material.

\begin{figure}[!ht]
    \centering
    \includegraphics[width=320pt]{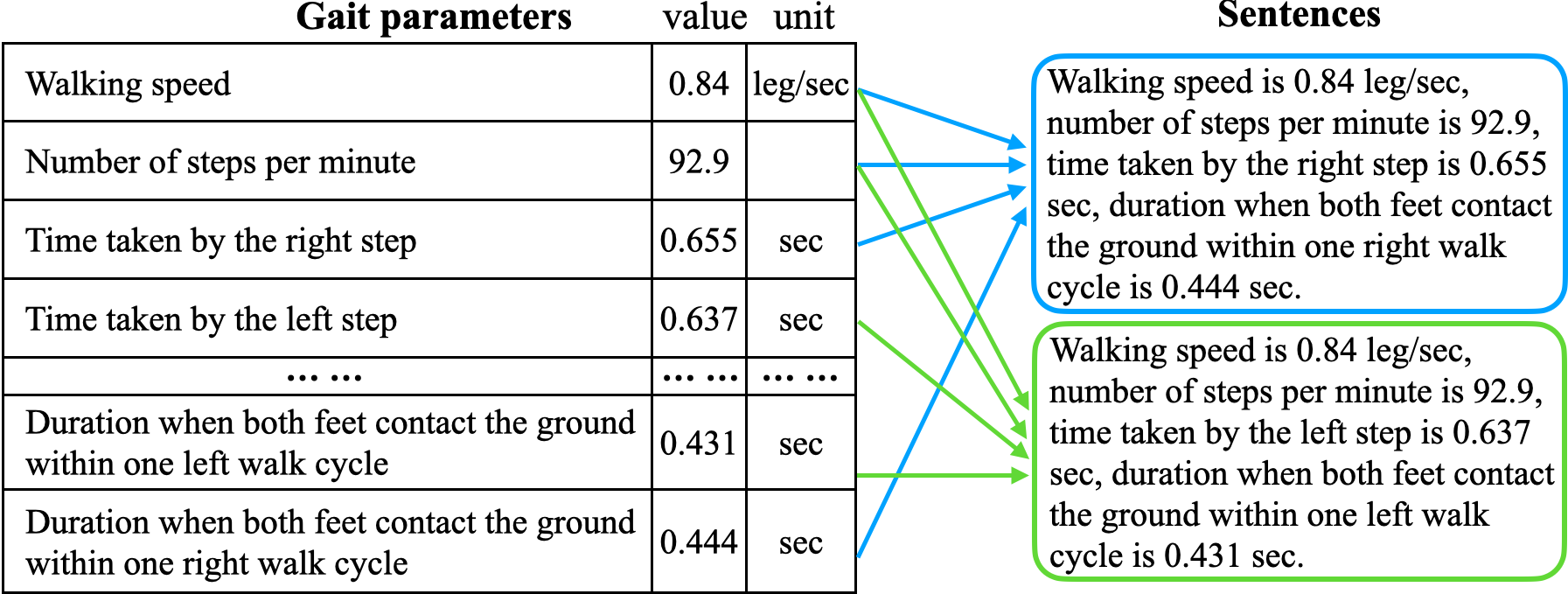}
    \caption{Translation of gait parameters into text.}
    \label{fig:gait params to sentence}
\end{figure}

\vspace{3pt}
\noindent\textbf{Preprocessing. }We crop the original videos based on bounding boxes, and employ a sliding window scheme (window size: 70 frames) to generate subsequences, with a stride of 25 for training and 0 for validation.
This process results in approximately 900 clips of 70 frames for each cross-validation fold.
To effectively incorporate the gait parameters into text space, we formulate sentences by combining four gait parameters with ``and", connecting names and values with ``is", as illustrated in Fig.\ref{fig:gait params to sentence}. The choice of four parameters per sentence is based on our observation that, in practice, neurologists often label a video by using only a few prominent or representative visual clues rather than exhaustively listing all evidences. Out of the total 29 parameters available, we select 438 combinations, each containing 4 parameters whose Pearson correlation coefficients are within the range of $[-0.4, 0.4]$. 

\subsection{VLM fine-tuning with visual and knowledge-aware prompts}
\label{sec: kapt}
\vspace{-1mm}
We adopt the prompt learning strategy, keeping the pre-trained VLM frozen to preserve its general representation and leverage the pre-aligned multi-modal latent space. 
Taking inspiration from KAPT \cite{kapt2023}, we introduce gait-specific knowledge-based prompts by feeding per-class descriptions ${Desc_i}$ (See our project page) into the text prompts.
These \textit{clinical gait notions} have been generated using ChatGPT-4 \cite{2023gpt4}, then subsequently filtered, modified, and validated by a neurologist.
To devise learnable prompts, we use KEPLER \cite{2021kepler}, similar to \cite{kapt2023}, on the class descriptions, which are then projected through per-class multi-layer perceptrons (MLPs), and added to the learnable parameters $\{X^k_i\}$ to form learnable prompts $\{C^k_i\}$:
\vspace{-3pt}
\begin{equation}
\vspace{-3pt}
    \{C^k_i\}_{i=1,...,N_{cls}} = Proj^k_{\phi}(\textit{KEPLER}(\{Desc_i\})) + \{X^k_i\},\quad k=1,...,8
\end{equation}
where $N_{cls}$ is the number of class, $C^k_i\in\mathbb{R}^{512}$ and $X^k_i\in\mathbb{R}^{512}$  represent the $k$-th learnable prompts and parameters associated to the $i$-th class, respectively. 
For the automatic prompt $\{D_i\}$, we extract keywords from $\{Desc_i\}$, as illustrated in Fig.\ref{fig: architecture}. More examples can be found on the project page. These selected texts then undergo standard tokenization of the frozen CLIP text encoder $\textit{FCLIP}_T$ to obtain $\{D_i\}$. Similarly, we pass the class names $\{T_i\}$ into the tokenizer of $\textit{FCLIP}_T$ to generate the class token $tok_i^{cls}$.
As shown in Fig.\ref{fig: architecture}, we concatenate $\{C_i\}$, $\{D_i\}$ and $\{tok_i^{cls}\}$ into  $\textit{FCLIP}_T$ to obtain the text features $\{F^T_i\}$:
\vspace{-3pt}
\begin{equation}
\vspace{-3pt}
    \{F_i^T\}=\textit{FCLIP}_T([\{C_i\},\{D_i\},\{tok_i^{cls}\}]).
\vspace{-3pt}
\end{equation}
On the video side, each frame of the input video $V$ goes through the tokenization of the Vision transformer (ViT) \cite{2020vit} and forms a sequence of per-frame representations $z_t^{(0)}$. 
The visual prompts for the $l$-th layer of the pretrained CLIP Vision Encoder $\textit{FCLIP}_V$ are derived by applying Vita-CLIP \cite{vita2023}'s video prompt learner 
($VitaVPL$)
to the output of the previous layer $\{z_t^{(l-1)}\}$:
\vspace{-3pt}
\begin{equation}
\vspace{-3pt}
    [S^{(l)}, G^{(l)}, L^{(l)}]_{l=1,...,12} = \textit{VitaVPL}_{\theta}(\{z_t^{(l-1)}\}),
\end{equation}
where $S^{(l)}$, $G^{(l)}$, and $L^{(l)}$ respectively denote the learnable summary, global, and local prompt tokens at layer $l$. 
As suggested in \cite{vita2023}, these prompt tokens are appended to $\{z_t^{(l-1)}\}$ and subsequently fed into $\textit{FCLIP}_V$ to obtain $F^V$:
\vspace{-5pt}
\begin{equation}
    F^V = \textit{FCLIP}_V([\{z_t^{(l-1)}\}, S^{(l)}, G^{(l)}, L^{(l)}]).
\vspace{-10pt}
\end{equation}
\\\noindent Moreover, to combat class imbalance, we employ a multi-class focal loss \cite{lu2020miccai} to maximize the cosine similarity of positive pairs:
\vspace{-5pt}
\begin{equation}
    L_{k} = \sum_{i=1}^{N_{cls}}-\alpha(1-p_i)^{\gamma}y_ilog(p_i),\quad p_i=\frac{exp(<F^T_i|F^V>/\tau)}{\sum^{N_{cls}}_{j=1}exp(<F^T_j|F^V>/\tau)},
\vspace{-5pt}
\end{equation}
where $y$ denotes the one-hot encoded label, $<\cdot|\cdot>$ the cosine similarity, and $\tau=0.01$ temperature parameter. We set the weighting factor $\alpha=0.25$ and the focusing parameter $\gamma=2$.
\vspace{-2mm}
\noindent\subsection{Contrastive learning with numerical text embeddings}
\label{sec: num gait text}
\vspace{-5mm}
\begin{figure}[!ht]
    \centering
    \includegraphics[width=250pt]{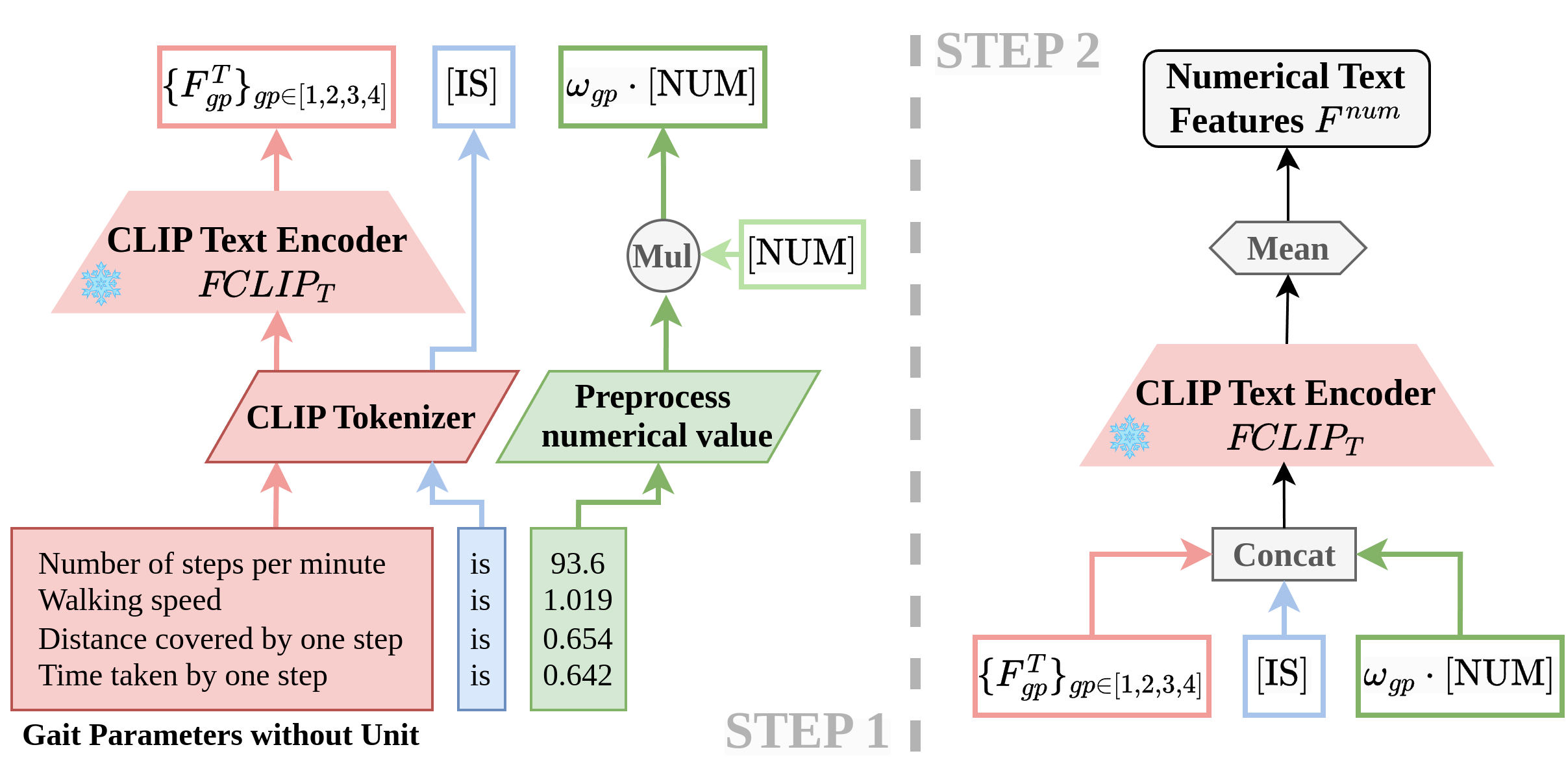}
    \caption{Our numerical text embedding (NTE) paradigm.}
    \label{fig:num text embed}
\vspace{-3mm}
\end{figure}
\noindent\textbf{Text embedding of numerical gait parameters.}
Starting from the set of sentences each containing four gait parameters, we employ a two-step embedding process as illustrated in Fig.\ref{fig:num text embed}. Initially, sentences without numerical values are fed into the CLIP text encoder, resulting in a descriptive embedding of the textual content $\{F_{gp}^T\}$. As illustrated in Fig.\ref{fig:num text embed}, we treat separately the logical conjunction ``is" to generate text embedding $\textbf{[IS]}$. Subsequently, number embeddings are generated by multiplying the dedicated embedding base $\textbf{[NUM]}$ with the associated numerical values $\{\omega_{gp}\}$. The chosen specialized embedding base is designed to be orthogonal to the position encoding \cite{2023xval}, ensuring the efficient transmission of numerical information through the self-attention blocks of the Transformer. The numerical text embedding $F^{num}$ is then obtained by applying the $\textit{FCLIP}_T$ to the concatenated sentence:
\vspace{-3pt}
\begin{equation}
\vspace{-3pt}
    F^{num} = \textit{FCLIP}_T(\{[ F^T_{gp},\textbf{[IS]},  \omega_{gp}\cdot\textbf{[NUM]} ]\}),\quad gp\in \{1,2,3,4\}.
\end{equation}

\begin{figure}[!ht]
    \centering
    \begin{subfigure}[b]{135pt}\includegraphics[width=135pt]{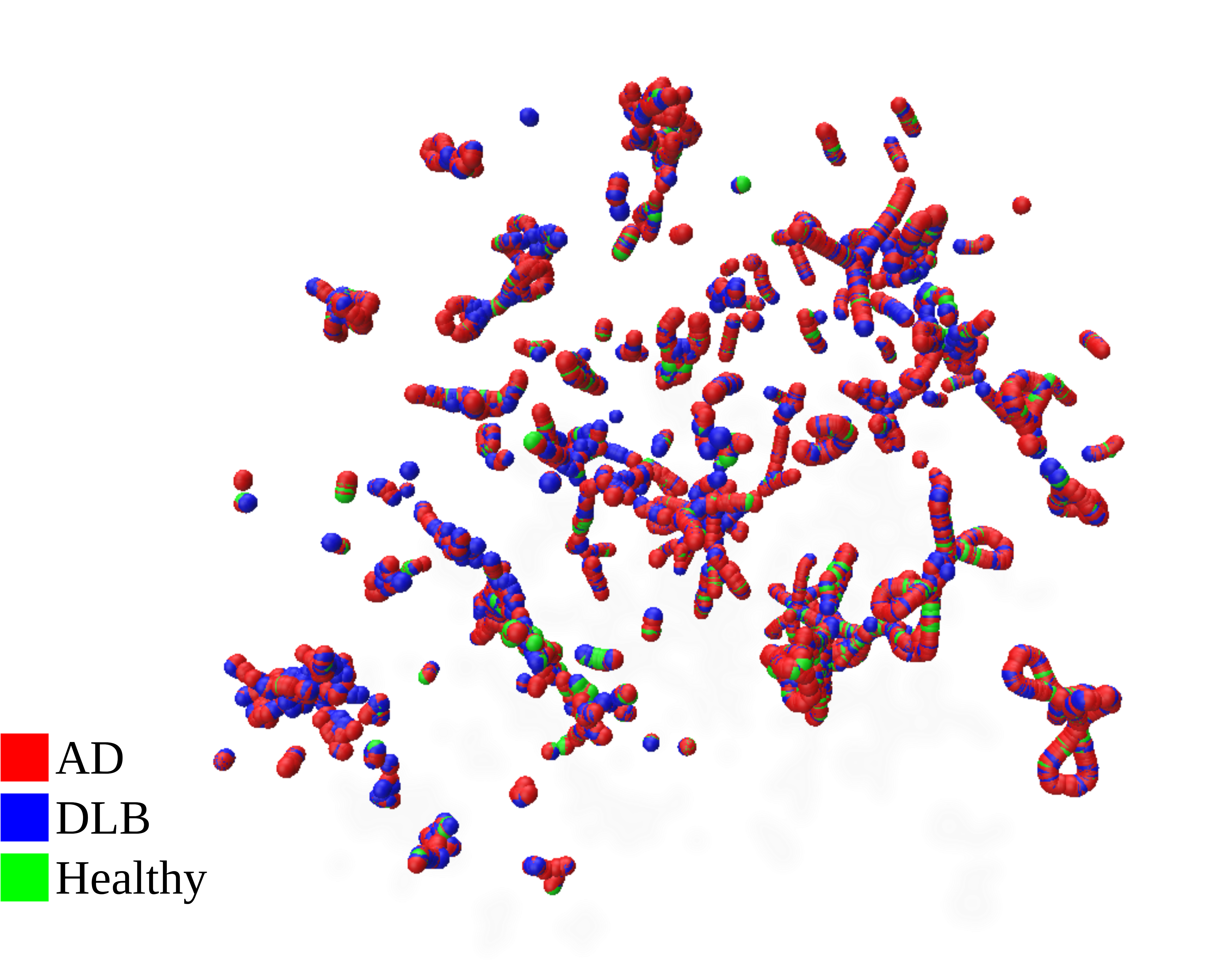}
    \caption{\centering Original text embeddings.}
    \end{subfigure}\hspace{5pt}
    \begin{subfigure}[b]{150pt}\includegraphics[width=135pt]{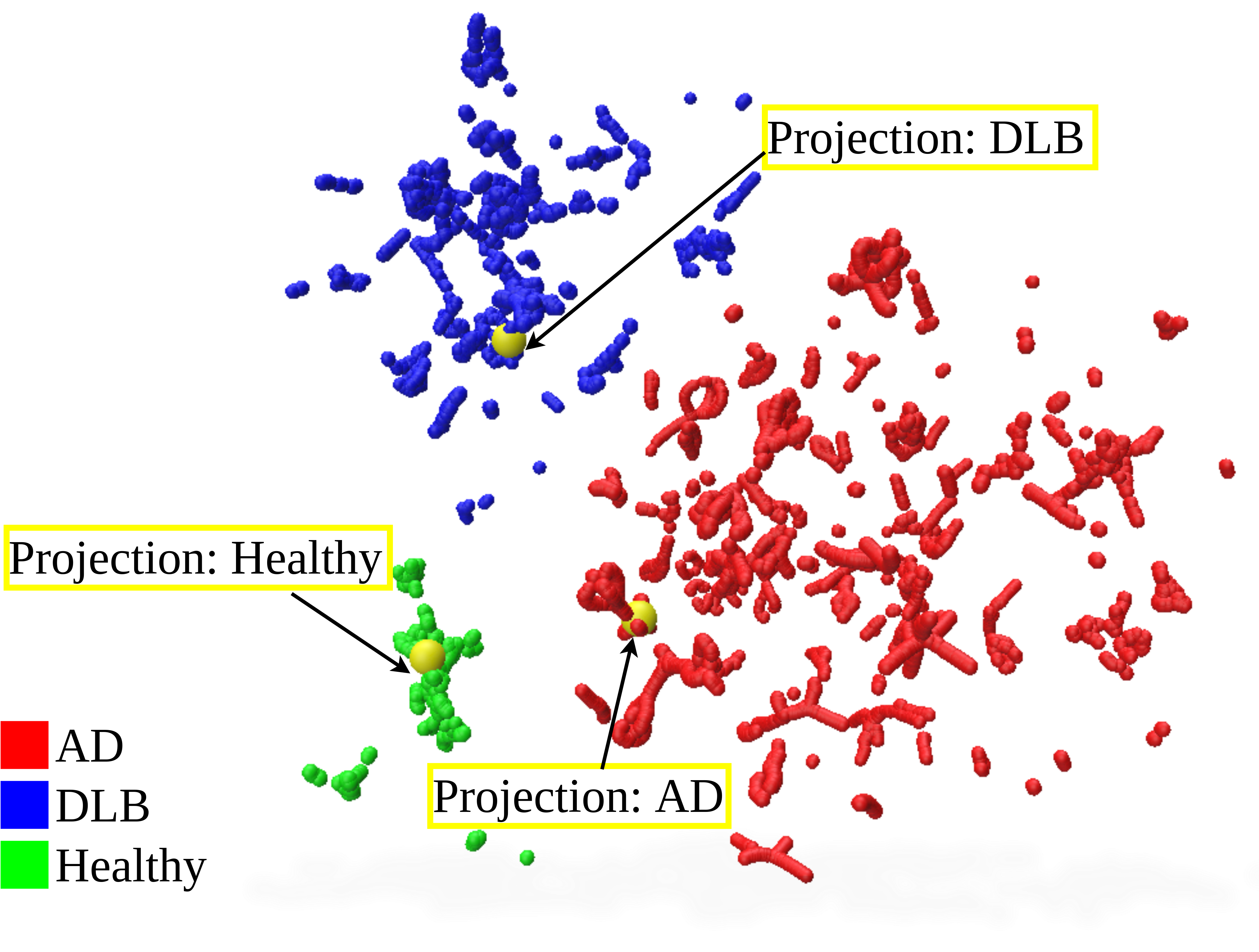}
    \caption{\centering Embeddings projected by MLPs.}
    \end{subfigure}
    \caption{Feature visualization using UMAP (no. components$=$3) for numerical text embeddings derived from gait parameters. 
    Yellow points in (b) represent the projections of the learned per-class text features. Images rendered with \href{www.polyscope.run}{Polyscope}.}
    \label{fig:gait param embeds}
\end{figure}

Fig.\ref{fig: compare number embeddings} in Supplementary Material demonstrates the cosine similarities of $F^{num}$ embedded from the text: ``the walking speed is [value]", where [value] ranges from 0 to 200, employing different methods to represent the numerical values. Our numerical embedding scheme, in contrast to adopting position encoding or directly encode digit and numerical text with $\textit{FCLIP}_T$, produces continuous embeddings that best reflect the numerical domain. Given that most gait parameter values are positive, we designate the mean value among healthy controls as the zero reference:
$V_{norm} = \alpha\cdot\frac{(V - \overline{V}_{healthy})}{\sigma}$, where $\sigma$ is the variance of the gait parameter values, and $\alpha$ is the scaling factor to adjust the data range to [-2.5, 2.5], the dynamic range of layer normalization within the self-attention block. 
The numerical text embeddings for the dementia grouping task are visualized in Fig.\ref{fig:gait param embeds}.
\vspace{-5mm}
\vspace{-10pt}

\noindent\textbf{Cross-modal contrastive learning.} 
To better align the multimodal representation with our task, we exploit  all accessible modalities, incorporating them into cross-modal contrastive learning: 
gait videos, class-specific descriptions, and numerical gait parameters. In our dataset, although gait videos are not always consistently paired with a corresponding gait parameter set, each set of gait parameters is linked to a video is assigned a class label. To this end, we transform these gait parameters into numerical text embeddings using the encoding method described earlier, and introduce classification tasks thereon.\\
\indent As illustrated in Fig.\ref{fig: architecture}, the projection of the generated text features ${F^T_i}$ and that of numerical embedding $F^{num}$ are trained in a way that the cosine similarity between $P^{num}$ and the projected text feature of its ground-truth class $P^T$ is maximized, by using a cross-entropy objective $L_{gp}$.
The global loss function becomes: $L = L_k + \omega\cdot L_{gp}$. We set $\omega=0.05$ through heuristic analysis. 
To demonstrate alignment of numerical embeddings with the multi-modal space, we visualize the embedding spaces before and after learning in Fig.\ref{fig:gait param embeds}.

\vspace{3pt}

\noindent\textbf{Interpreting the per-class text embedding.} We aim at the effective translation of per-class text features $\{F^T_{i}\}$ into natural language expressions using the vocabulary of numerical gait parameters. To this end, we trained a text decoder from scratch to transform numerical text embeddings $F^{num}$ back into their corresponding gait parameters, reverting the embedding scheme shown in Fig.\ref{fig:num text embed}. An 4-layer transformer decoder $\textit{D}_T$ is employed for text decoding. 
In line with recent developments in text-only decoder pre-training \cite{2023decap}, we train $\textit{D}_T$ using the prefix language modeling. Specifically, given a sentence composed of gait parameters $\textbf{s} = \{\text{word}_1,\text{word}_2, ..., \text{word}_L\}$, we generate a sequence of token IDs using the dictionary of $\textit{FCLIP}_T$. For the numbers, 
we \textit{scale} the numbers, which had been previously normalized to [-2.5, 2.5], to a graduated integer scale of [0,$N_{num}$]. The token ID $tok$ of a number [$\textbf{num}$] is defined as: $tok=[\textbf{EOS}]+\textit{scale}([\textbf{num}])$, where $[\textbf{EOS}]=49407$. $\textit{D}_T$ learns to reconstruct the sequence of token IDs $\{tok_j\}$ starting from the numerical text embedding $F^{num}$.
In addition to the vanilla cross-entropy loss \cite{2023decap}, we leverage an ordinal cross-entropy loss to further penalize the reconstruction error of the number values:
\vspace{-6pt}
\begin{equation}
    L_{num}=-\frac{|\hat{tok}-tok|}{[\textbf{EOS}]+N_{num}-1}\sum^{[\textbf{EOS}]+N_{num}}_{m=1}y_{m}log(p_m),
\vspace{-7pt}
\end{equation}
where $N_{num}=200$, $|\hat{tok}-tok|$ represents the absolute distance between the ground-truth token ID $tok$ and the estimation $\hat{tok}$, $y$ denotes the one-hot encoded ground-truth label, and $p$ the estimated probability. \\
\indent Benefiting from the proposed cross-modal contrastive learning scheme, $\{F^T_{i}\}$ can be represented as a linear combination of the numerical text embeddings ${F^{num}}$, with weights computed by measuring the cosine similarity between $\{P_i\}$ and $P^{num}$.
Subsequently, we apply $\textit{D}_T$ on $\{\hat{F}^{num}_i\}$ to generate natural language descriptions: $\{\hat{\textit{Desc}}_i\}=\mathbf{D}^T(\{\hat{F}^{num}_i\})$.
\vspace{-7pt}
\section{Experiments and Results}
\vspace{-6pt}
Our study includes two classification tests:\textit{ Gait scoring} to estimate the severity of a patient's condition based on a 4-class gait scoring (normal--0, slight--1, mild--2, and moderate--3) following MDS-UPDRS III \cite{mds-updrs2008}, and \textit{dementia subtyping} to distinguish between different dementia groups: normal/DLB(Dementia with Lewy Bodies)/AD(Alzheimer’s Disease). See the project page for detailed clinical gait descriptions on each class. 
Due to its limited size (a total of 118 videos), we divide our video dataset into training and validation sets and conduct 10-fold cross-validation for each classification task. 
Confusion matrices are provided in Fig.\ref{fig:fig_confmat} of Supplementary Material.
\vspace{-7mm}
\begin{table}
\caption{Comparative analysis on two classification tasks: Gait score (`Gait scoring') and dementia subtyping (`Dem. group'). Model performance is evaluated using top-1 accuracy (`acc',\%) and F1-score (`Fscore',\%).
\vspace{5pt}
}
\label{tab:ablation studies+sota}
\begin{minipage}{.51\textwidth} 
\centering

\textbf{(a) Different model configurations}
{\scriptsize
\begin{tabu}to 0.95\textwidth{|X[l, 3]|X[l, 1]|X[l, 1]|X[l, 1]|X[l, 1]|}
\hline
Model configurations&\multicolumn{2}{l|}{Gait scoring}&\multicolumn{2}{l|}{Dem. group}\\
\cline{2-5}&Acc. &Fscore &Acc. &Fscore \\
\hline
Baseline&64.78 &60.75 &86.27 &79.24 \\
Baseline\small{+KAPT} &65.98 &61.97 &87.29 &78.48\\
Baseline\small{+NTE} &64.44 &57.64 &88.26 &81.34 \\
Ours&\textbf{67.76} &\textbf{62.59} &\textbf{90.08} &\textbf{83.86}\\
\hline
\end{tabu}}
\end{minipage}%
\begin{minipage}{.5\textwidth} 
\centering
\textbf{(b) SOTA methods}
{\scriptsize
\begin{tabu}to 0.95\textwidth{|X[l, 3]|X[l, 1]|X[l, 1]|X[l, 1]|X[l, 1]|}
\hline
State-of-the-art models&\multicolumn{2}{l|}{Gait scoring}&\multicolumn{2}{l|}{Dem. group}\\
\cline{2-5}&Acc. &Fscore &Acc. &Fscore \\
\hline
OF-DDNet\cite{lu2020miccai} &54.73 &48.59 &68.92 &65.38 \\
ST-GCN \cite{sabo2022estimating} &49.08 &43.87 &61.46 &56.99 \\
KShapeNet\cite{friji2021geometric} &53.69 &44.85 &65.27 &54.86\\
GaitBase\cite{opengait2023} &43.48 &30.25 &53.42 &41.76\\
\hline
\end{tabu}}
\end{minipage}

\end{table}
\subsection{Ablation studies} 
\label{sec: ablation study}
We conduct ablation experiments on four model configurations, with different combinations of  knowledge augmented prompt tuning (KAPT) and numerical text embedding (NTE).
As shown in Table \ref{tab:ablation studies+sota}(a), the combination of both KAPT and NTE yield the best performance,  
whereas NTE alone or KAPT alone can sometimes slightly worsen the performance.
In general, the models tend to perform better for dementia group task, which can be attributed to the more distinctive per-class descriptions and more objective ground truth labeling in that classification.

\vspace{-2mm}
\subsection{Comparison with state-of-the-art}
\label{sec: SOTA}
\vspace{-1mm}
We compare our model with several related state-of-the-art (SOTA) models. Three of these models (\cite{lu2020miccai,sabo2022estimating,max-gr2023}) are specifically designed for the classification of Parkinsonism severity on 3D skeletons, whereas GaitBase \cite{opengait2023} is for the gait recognition and silhouette-based. As shown in Table \ref{tab:ablation studies+sota}, our method achieves the best overall results in both tasks. This is somewhat expected, as other models are not designed for, and do not adapt well to, the constraints of limited data size. 
Differing from other models where the performance is averaged across 10 folds, result of GaitBase\cite{opengait2023} represents the best across 5 folds.
Note that the chosen SOTA methods operate on 3D reconstructed poses or extracted 2D silhouettes, for which we used VIBE\cite{VIBE}, MAX-GRNet\cite{max-gr2023} or PARE\cite{2021pare}. Changing among these algorithms did not result in meaningful differences in performance. The results in Table \ref{tab:ablation studies+sota}(b) represent the best among these combinations. 

\subsection{Decoding the per-class description}
\label{decode F^T}
\vspace{-1mm}
We apply the pretrained text decoder $\textit{D}_T$ on the per-class text features $\{F_i\}$ obtained through the cross-modal contrastive learning in Sec.\ref{sec: num gait text}. Examples of the decoded texts are shown in Fig.\ref{fig:decoded text features} of Supplementary Material and in the accompanying video. For the gait scoring task, descriptions of \textit{slight} impairment bear a closer similarity to \textit{normal} than to \textit{mild} or \textit{moderate} impairments, conforming to the severity levels. Most notably, descriptions on \textit{moderate} manifest significant abnormalities, including marked deviation of the foot angle from the progression line. 
Additionally, we observe that certain criteria in the clinical gait notions have been mapped to some quantitative gait parameters, such as the slowness in \textit{mild} impairment.
For the dementia subtype task, the decoded sentences for the DLB clearly show that the model has learned the distinctive gait characteristics. 
This aligns with findings in \cite{medical2020differentiating}, where the motor symptom in the DLB group is greater compared to the AD group.
The decoded texts for the AD group, on the other hand, are less distinctive and are rather close to the healthy group. 
This may be attributed to the limited availability, with only 6 healthy elderly videos having corresponding gait parameters during training.

\vspace{-2mm}
\section{Conclusion}
\vspace{-2mm}
We presented a knowledge augmentation strategy to enhance the adaptability of a large-scale pre-trained Vision-Language Model for video-based gait analysis in neurodegenerative diseases.
Our method makes use of class-specific descriptive text and numerical gait parameters associated with patient videos, via prompt learning, and numeracy-enhanced text representation, respectively.
On two video-based gait classifications tasks, our model significantly outperformed other strong SOTA methods, given only slightly more than 100 videos.
We believe that our work demonstrates how to efficiently enhance representation learning and offers a novel alternative to incorporating patient metadata, particularly in tabular form.

\subsection*{Acknowledgements}
This work has been funded by the French national project “ArtIC” (Artificial Intelligence for Care, ANR-20-THIA-0006) and the binational project “Synthetic Data Generation and Sim-to-Real Adaptive Learning for Real-World Human Daily Activity Recognition of Human-Care Robots (21YS2900)” granted by the ETRI, South Korea.
Kun Yuan has been supported by the European Union project “CompSURG” (ERC, 101088553).

\bibliographystyle{splncs04}\bibliography{ref}

\begin{thebibliography}{10}
\providecommand{\url}[1]{\texttt{#1}}
\providecommand{\urlprefix}{URL }
\providecommand{\doi}[1]{https://doi.org/#1}

\bibitem{2023gpt4}
Achiam, J., Adler, S., Agarwal, S., Ahmad, L., Akkaya, I., Aleman, F.L., Almeida, D., Altenschmidt, J., Altman, S., Anadkat, S., et~al.: Gpt-4 technical report. arXiv preprint arXiv:2303.08774  (2023)

\bibitem{St-LSTMGaitClassify}
Albuquerque, P., Verlekar, T.T., Correia, P.L., Soares, L.D.: A spatiotemporal deep learning approach for automatic pathological gait classification. Sensors  \textbf{21}(18), ~6202 (2021)

\bibitem{2020vit}
Dosovitskiy, A., Beyer, L., Kolesnikov, A., Weissenborn, D., Zhai, X., Unterthiner, T., Dehghani, M., Minderer, M., Heigold, G., Gelly, S., et~al.: An image is worth 16x16 words: Transformers for image recognition at scale. In: Conference on Neural Information Processing Systems (NeurIPS) (2020)

\bibitem{opengait2023}
Fan, C., Liang, J., Shen, C., Hou, S., Huang, Y., Yu, S.: Opengait: Revisiting gait recognition towards better practicality. In: Proceedings of the IEEE/CVF Conference on Computer Vision and Pattern Recognition. pp. 9707--9716 (2023)

\bibitem{friji2021geometric}
Friji, R., Drira, H., Chaieb, F., Kchok, H., Kurtek, S.: Geometric deep neural network using rigid and non-rigid transformations for human action recognition. In: Proceedings of the IEEE/CVF international conference on computer vision. pp. 12611--12620 (2021)

\bibitem{mds-updrs2008}
Goetz, C.G., Tilley, B.C., Shaftman, S.R., Stebbins, G.T., Fahn, S., Martinez-Martin, P., Poewe, W., Sampaio, C., Stern, M.B., Dodel, R., et~al.: Movement disorder society-sponsored revision of the unified parkinson's disease rating scale (mds-updrs): scale presentation and clinimetric testing results. Movement disorders: official journal of the Movement Disorder Society  \textbf{23}(15),  2129--2170 (2008)

\bibitem{2023xval}
Golkar, S., Pettee, M., Eickenberg, M., Bietti, A., Cranmer, M., Krawezik, G., Lanusse, F., McCabe, M., Ohana, R., Parker, L., et~al.: xval: A continuous number encoding for large language models. arXiv preprint arXiv:2310.02989  (2023)

\bibitem{huang2021gloria}
Huang, S.C., Shen, L., Lungren, M.P., Yeung, S.: Gloria: A multimodal global-local representation learning framework for label-efficient medical image recognition. In: Proceedings of the IEEE/CVF International Conference on Computer Vision. pp. 3942--3951 (2021)

\bibitem{huang2023visual}
Huang, Z., Bianchi, F., Yuksekgonul, M., Montine, T.J., Zou, J.: A visual--language foundation model for pathology image analysis using medical twitter. Nature medicine  \textbf{29}(9),  2307--2316 (2023)

\bibitem{kapt2023}
Kan, B., Wang, T., Lu, W., Zhen, X., Guan, W., Zheng, F.: Knowledge-aware prompt tuning for generalizable vision-language models. In: Proceedings of the IEEE/CVF International Conference on Computer Vision. pp. 15670--15680 (2023)

\bibitem{VIBE}
Kocabas, M., Athanasiou, N., Black, M.J.: Vibe: Video inference for human body pose and shape estimation. In: Proceedings of the IEEE/CVF conference on computer vision and pattern recognition. pp. 5253--5263 (2020)

\bibitem{2021pare}
Kocabas, M., Huang, C.H.P., Hilliges, O., Black, M.J.: Pare: Part attention regressor for 3d human body estimation. In: Proceedings of the IEEE/CVF International Conference on Computer Vision. pp. 11127--11137 (2021)

\bibitem{2023decap}
Li, W., Zhu, L., Wen, L., Yang, Y.: Decap: Decoding clip latents for zero-shot captioning via text-only training. arXiv preprint arXiv:2303.03032  (2023)

\bibitem{lu2020miccai}
Lu, M., Poston, K., Pfefferbaum, A., Sullivan, E.V., Fei-Fei, L., Pohl, K.M., Niebles, J.C., Adeli, E.: Vision-based estimation of mds-updrs gait scores for assessing parkinson’s disease motor severity. In: International Conference on Medical Image Computing and Computer-Assisted Intervention. pp. 637--647. Springer (2020)

\bibitem{ImpactofEnvironment}
Mc~Ardle, R., Del~Din, S., Donaghy, P., Galna, B., Thomas, A.J., Rochester, L.: The impact of environment on gait assessment: considerations from real-world gait analysis in dementia subtypes. Sensors  \textbf{21}(3), ~813 (2021)

\bibitem{medical2020differentiating}
Mc~Ardle, R., Del~Din, S., Galna, B., Thomas, A., Rochester, L.: Differentiating dementia disease subtypes with gait analysis: feasibility of wearable sensors? Gait \& posture  \textbf{76},  372--376 (2020)

\bibitem{pathologicalSignature}
Mc~Ardle, R., Galna, B., Donaghy, P., Thomas, A., Rochester, L.: Do alzheimer's and lewy body disease have discrete pathological signatures of gait? Alzheimer's \& Dementia  \textbf{15}(10),  1367--1377 (2019)

\bibitem{toaw2022}
Mehdizadeh, S., Nabavi, H., Sabo, A., Arora, T., Iaboni, A., Taati, B.: The toronto older adults gait archive: video and 3d inertial motion capture data of older adults’ walking. Scientific data  \textbf{9}(1), ~398 (2022)

\bibitem{QuantativeGA}
Merory, J., Wittwer, J., Rowe, C., Webster, K.: Quantitative gait analysis in patients with dementia with lewy bodies and alzheimer's disease. Gait \& posture  \textbf{26},  414--9 (10 2007). \doi{10.1016/j.gaitpost.2006.10.006}

\bibitem{miech2020end}
Miech, A., Alayrac, J.B., Smaira, L., Laptev, I., Sivic, J., Zisserman, A.: End-to-end learning of visual representations from uncurated instructional videos. In: Proceedings of the IEEE/CVF Conference on Computer Vision and Pattern Recognition. pp. 9879--9889 (2020)

\bibitem{muller2018correlation}
Muller, C., Perisse, J., Blanc, F., Kiesmann, M., Astier, C., Vogel, T.: Corr{\'e}lation des troubles de la marche au profil neuropsychologique chez les patients atteints de maladie d’alzheimer et maladie {\`a} corps de lewy. Revue Neurologique  \textbf{174},  S2--S3 (2018)

\bibitem{qin2022medical}
Qin, Z., Yi, H., Lao, Q., Li, K.: Medical image understanding with pretrained vision language models: A comprehensive study. arXiv preprint arXiv:2209.15517  (2022)

\bibitem{clip2021}
Radford, A., Kim, J.W., Hallacy, C., Ramesh, A., Goh, G., Agarwal, S., Sastry, G., Askell, A., Mishkin, P., Clark, J., et~al.: Learning transferable visual models from natural language supervision. In: International conference on machine learning. pp. 8748--8763. PMLR (2021)

\bibitem{sabo2022estimating}
Sabo, A., Mehdizadeh, S., Iaboni, A., Taati, B.: Estimating parkinsonism severity in natural gait videos of older adults with dementia. IEEE journal of biomedical and health informatics  \textbf{26}(5),  2288--2298 (2022)

\bibitem{max-gr2023}
Wang, D., Zouaoui, C., Jang, J., Drira, H., Seo, H.: Video-based gait analysis for assessing alzheimer's disease and dementia with lewy bodies. In: Wu, S., Shabestari, B., Xing, L. (eds.) Applications of Medical Artificial Intelligence. pp. 72--82. Springer Nature Switzerland, Cham (2024)

\bibitem{2021kepler}
Wang, X., Gao, T., Zhu, Z., Zhang, Z., Liu, Z., Li, J., Tang, J.: Kepler: A unified model for knowledge embedding and pre-trained language representation. Transactions of the Association for Computational Linguistics  \textbf{9},  176--194 (2021)

\bibitem{wang2022medclip}
Wang, Z., Wu, Z., Agarwal, D., Sun, J.: Medclip: Contrastive learning from unpaired medical images and text. In: Proceedings of the 2022 Conference on Empirical Methods in Natural Language Processing. pp. 3876--3887 (2022)

\bibitem{vita2023}
Wasim, S.T., Naseer, M., Khan, S., Khan, F.S., Shah, M.: Vita-clip: Video and text adaptive clip via multimodal prompting. In: Proceedings of the IEEE/CVF Conference on Computer Vision and Pattern Recognition. pp. 23034--23044 (2023)

\bibitem{yuan2023learning}
Yuan, K., Srivastav, V., Yu, T., Lavanchy, J., Mascagni, P., Navab, N., Padoy, N.: Learning multi-modal representations by watching hundreds of surgical video lectures. arXiv preprint arXiv:2307.15220  (2023)

\end{thebibliography}
\newpage
\section{Supplementary Material}
\vspace{-30pt}
\setcounter{figure}{0}
\setcounter{table}{0}
\begin{figure}
    \begin{subfigure}[s]{90pt}
        \includegraphics[width=90pt]{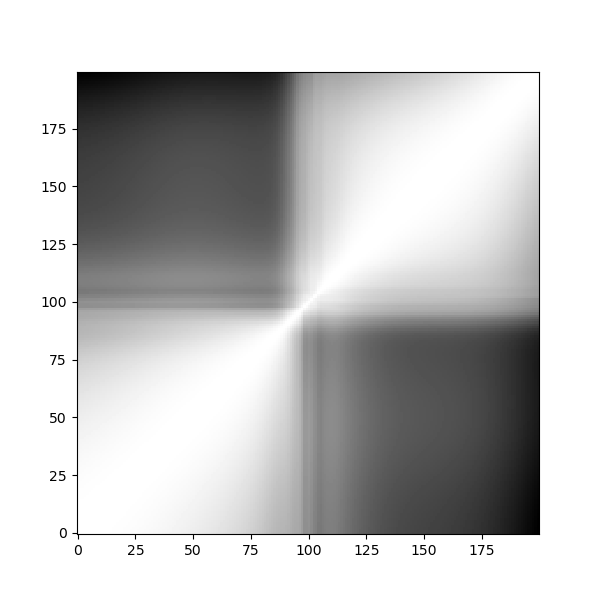}
        \vspace{-15pt}
        \caption{\centering Ours.}
    \end{subfigure}\hspace{-10pt}
    \begin{subfigure}[s]{90pt}
        \includegraphics[width=90pt]{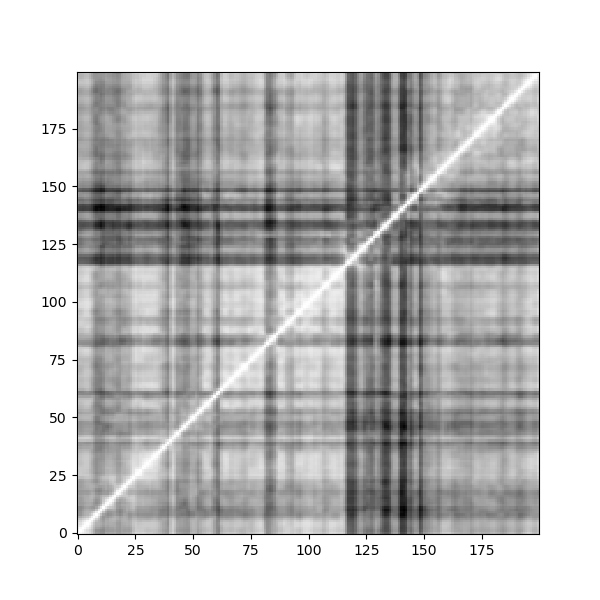}
        \vspace{-15pt}
        \caption{\centering Position encoding.}
    \end{subfigure}\hspace{-10pt}
    \begin{subfigure}[s]{90pt}
        \includegraphics[width=90pt]{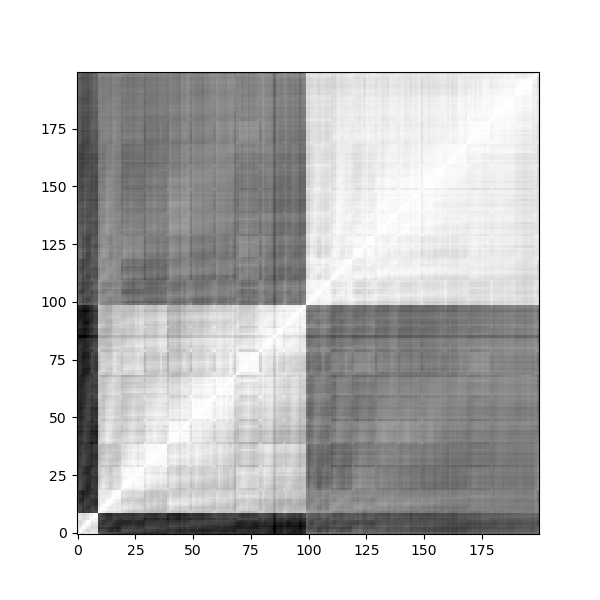}
        \vspace{-15pt}
        \caption{\centering Digits.}
    \end{subfigure}\hspace{-10pt}
    \begin{subfigure}[s]{90pt}
        \includegraphics[width=90pt]{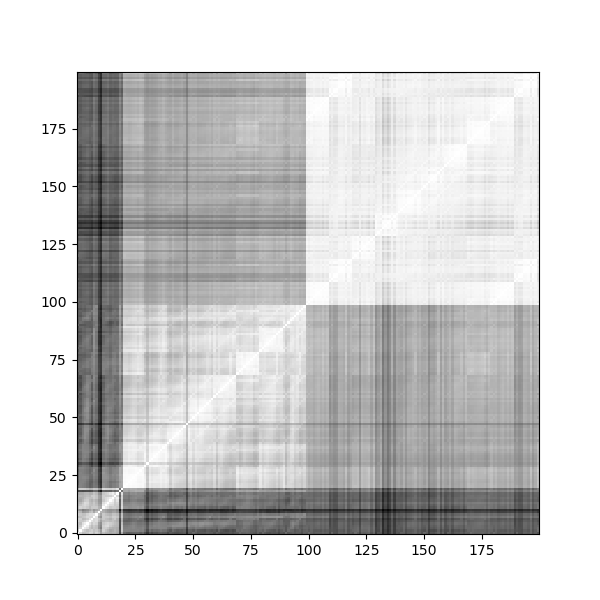}
        \vspace{-15pt}
        \caption{\centering Number texts.}
    \end{subfigure}
    \caption{Cosine similarity among text embeddings derived from gait parameters. The text used is ``the walking speed is [value]", where [value] ranges from 0 to 200. In comparison to others, our method produces a smooth, continuous similarity map across the value domain.}
    \label{fig: compare number embeddings}
\end{figure}

\noindent\begin{table}[!ht]
\vspace{-45pt}
\caption{Descriptions of 29 gait parameters used in our study.\vspace{4pt}}
\label{tab: 29 gait params}
\begin{tabu}{|l|l|}
\hline
ID&\multicolumn{1}{c|}{Gait parameter description} \\
\hline
1&Walking speed\\
2&Number of steps per minute\\
3&Difference in step duration between a left step and a right step\\
4&Difference in distance covered between a left step and a right step\\
5&Time difference between the left walk cycle and the right walk cycle\\
6&Time taken by the right step\\
7&Time taken by the left step\\
8&Distance covered by the right step\\
9&Distance covered by the left step\\
10&Time taken from the first contact of the left foot to the second contact of the left foot\\
11&Time taken from the first contact of the right foot to the second contact of the right foot\\
12&Distance covered by one left walk cycle\\
13&Distance covered by one right walk cycle\\
14&Vertical height covered from the heel point of left foot to the second contact of left foot\\
15&Vertical height covered from the heel point of right foot to the second contact of right foot\\
16&Percentage of the duration when the left foot is off the ground within one walk cycle\\
17&Percentage of the time when the right foot is off the ground within one walk cycle\\
18&Time when the left foot is off the ground within one walk cycle\\
19&Time when the right foot is off the ground within one walk cycle\\
20&Percentage of the duration when only the left foot contacts the ground within one walk cycle\\
21&Percentage of the duration when only the right foot contacts the ground within one walk cycle\\
22&Duration when only the left foot contacts the ground within one walk cycle\\
23&Duration when only the right foot contacts the ground within one walk cycle\\
24&Percentage of the duration when both feet contact the ground within left walk cycle\\
25&Percentage of the time when both feet contact the ground within right walk cycle\\
26&Duration when both feet contact the ground within left walk cycle\\
27&Duration when both feet contact the ground within right walk cycle\\
28&Angle between the progression line of left foot and the line from left heel to forefoot pressure center\\
29&Angle between the progression line of right foot and the line from right heel to forefoot pressure center\\
\hline
\end{tabu}
\end{table}
\vspace{-10mm}
\begin{figure}[!ht]
    \centering
    \begin{subfigure}[c]{330pt}\includegraphics[width=330pt]{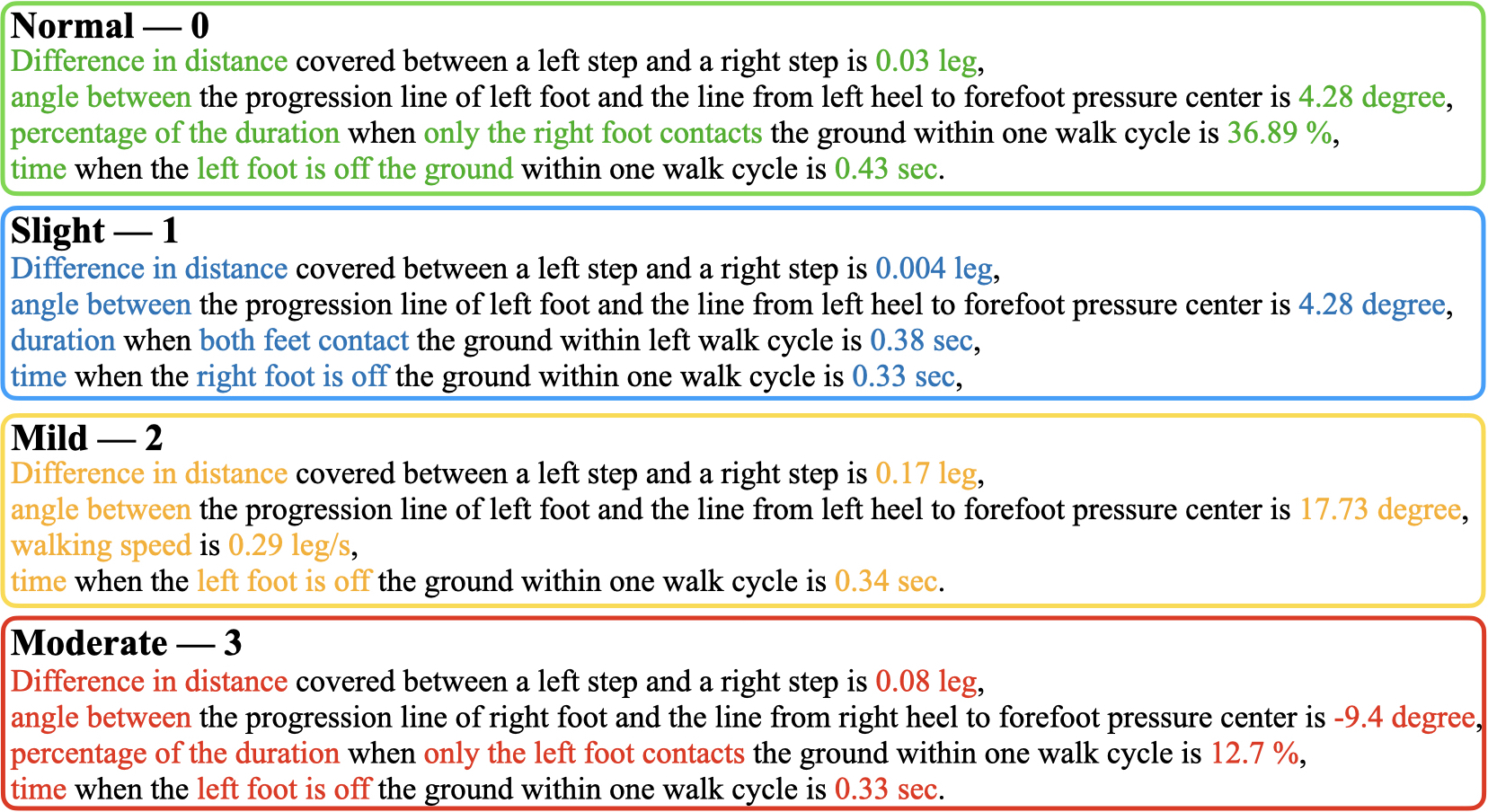}
    \caption{\centering MDS-UPDRS gait scores.}
    \end{subfigure}\\
    \begin{subfigure}[c]{330pt}\includegraphics[width=330pt]{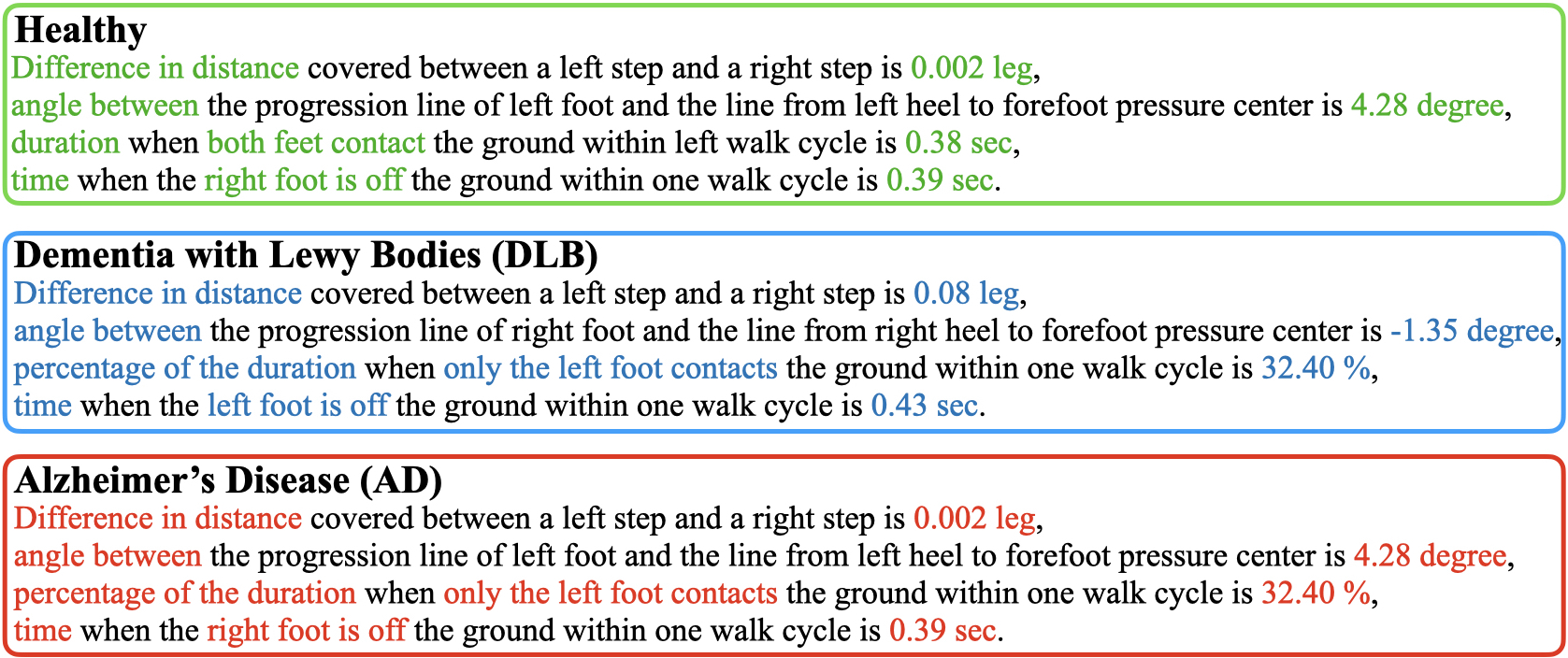}
    \caption{\centering Diagnostic groups.}
    \end{subfigure}
    \caption{ Descriptions generated from per-class text features through the pretrained text decoder. Key criteria are highlighted in the respective class color.}
    \label{fig:decoded text features}
\end{figure}

\begin{figure}[!htbp]
    \centering
    \begin{subfigure}[b]{170pt}\includegraphics[width=170pt]{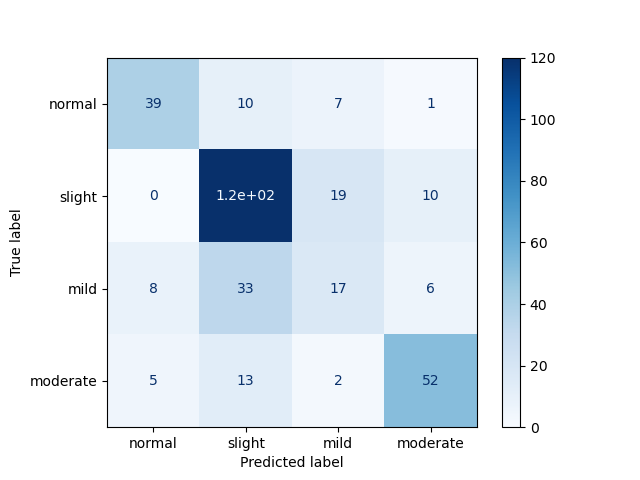}
    \caption{\centering Gait score classification.}
    \end{subfigure}
    \begin{subfigure}[b]{170pt}\includegraphics[width=170pt]{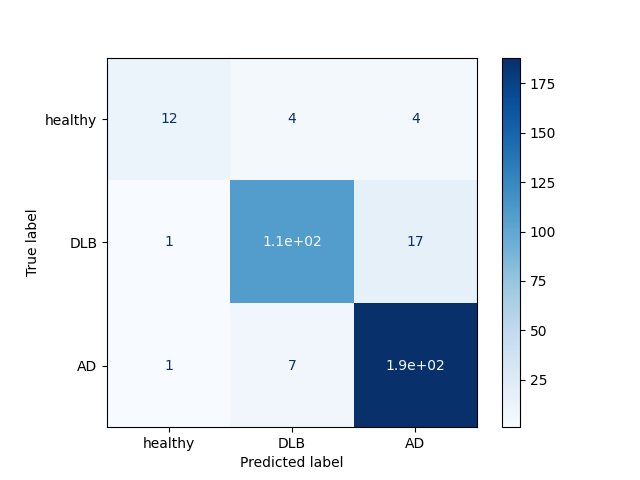}
    \caption{\centering Dementia subtype classification.}
    \end{subfigure}
    \caption{Confusion matrices for the classification tasks.}
    \label{fig:fig_confmat}
\end{figure}
\end{document}